\documentclass[conference, a4paper]{IEEEtran}
\IEEEoverridecommandlockouts
\usepackage{cite}
\usepackage{amsmath,amssymb,amsfonts}
\usepackage{algorithmic}
\usepackage{graphicx}
\usepackage{textcomp}
\usepackage{xcolor}
\usepackage{stfloats}
\graphicspath{{figs/}}
\usepackage{booktabs}
\usepackage{etoolbox}
\usepackage{siunitx}
\usepackage{textcomp}
\usepackage{multirow}
\usepackage{siunitx}
\usepackage{hyperref}
\usepackage{pifont}
\let\OLDthebibliography\thebibliography
\renewcommand\thebibliography[1]{
  \OLDthebibliography{#1}
  \setlength{\parskip}{0pt}
  \setlength{\itemsep}{0pt plus 0.3ex}
}

\begin{document}\sloppy

\def\x{{\mathbf x}}
\def\L{{\cal L}}

\title{Privacy-Preserving Image Classification Using ConvMixer with Adaptive Permutation Matrix
}

\author{\IEEEauthorblockN{Zheng Qi, AprilPyone MaungMaung and Hitoshi Kiya}
	\IEEEauthorblockA{Tokyo Metropolitan University, Asahigaoka, Hino-shi, Tokyo, 191--0065, Japan}
}

\maketitle

\begin{abstract}
In this paper, we propose a privacy-preserving image classification method using encrypted images under the use of the ConvMixer structure. Block-wise scrambled images, which are robust enough against various attacks, have been used for privacy-preserving image classification tasks, but the combined use of a classification network and an adaptation network is needed to reduce the influence of image encryption. However, images with a large size cannot be applied to the conventional method with an adaptation network because the adaptation network has so many parameters. Accordingly, we propose a novel method, which allows us not only to apply block-wise scrambled images to ConvMixer for both training and testing without the adaptation network, but also to provide a higher classification accuracy than conventional methods. 
\end{abstract}
\section{Introduction}
\label{sec:intro}
The spread of deep neural networks (DNNs)~\cite{p1} has greatly contributed to solving complex tasks for many applications. Recently, it has been very popular for data owners to utilize cloud servers to compute and process a large amount of data. However, data privacy may be compromised in that process, so it is necessary to protect data privacy in cloud environments, and privacy-preserving DNNs have become an urgent challenge~\cite{p9,p10}. Homomorphic encryption methods~\cite{h1,h2,h3,h4,h5,h6,h7,h8} may contribute to such a problem, but the computation and memory costs are high, and it is not easy to apply these methods to DNNs directly. 
In this paper, we focus on protecting data privacy by encrypting data before uploading the data to the cloud environment.

Privacy-preserving image classification methods have to satisfy two requirements: high classification accuracy and strong robustness against various attacks. Image encoding approaches for privacy-preserving image classification such as GAN-based method~\cite{p15} achieved a high classification accuracy, but are not robust against some attacks~\cite{p8}. On the other hand, the use of block-wise scrambled images has been confirmed to be robust against various attacks, but it is difficult to avoid the influence of the image encryption~\cite{p13,p14}. One of the solutions is to use a classification network with an adaptation network~\cite{p2,p3}. However, the adaptation network used for reducing the influence of encryption has so many parameters, so images with a large size cannot be applied to the classification network. To overcome the problem, we propose the combined use of a novel block-wise encryption method and a ConvMixer~\cite{p4} with an adaptive permutation matrix. In an experiment, the proposed method is confirmed to outperform conventional methods in terms of classification performance.

\section{Related work}
Tanaka first introduced a block-wise learnable image encryption (LE)~\cite{p2} method with an adaptation layer, which is used prior to a classifier to reduce the influence of image encryption. Another encryption method is a pixel-wise encryption (PE) method in which negative-positive transformation and color component shuffling are applied without using any adaptation layer~\cite{p11}. However, both encryption methods are not robust enough against ciphertext-only attacks as reported in~\cite{p8,p12}. To enhance the security of encryption, LE was extended by adding a block scrambling step and a pixel encryption operation with multiple keys (hereinafter denoted as ELE)~\cite{p3}. However, ELE still has a lower accuracy than that of using plain images, although an additional adaptation network is applied to reduce the influence of the encryption. Moreover, images with a large size cannot be applied to ELE because the use of the adaptation network increases the number of parameters in the model. Accordingly, we propose a novel privacy-preserving classification method to improve these issues that the conventional methods have.

\section{Proposed Privacy-Preserving Method}
\begin{figure*}[htbp]
	\centerline{\includegraphics[width=18cm]{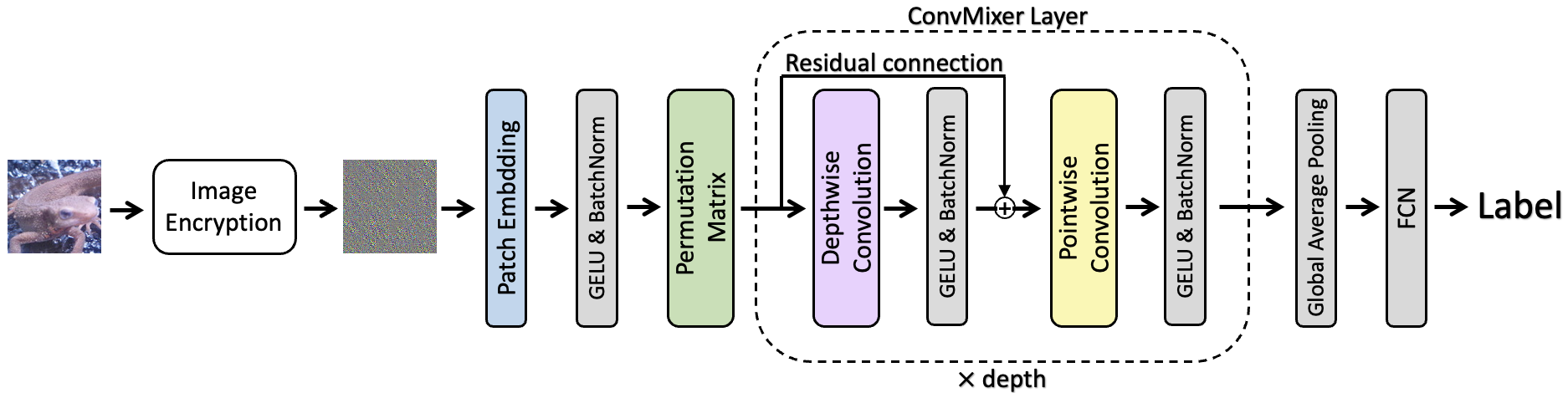}}
	\caption{Framework of proposed method with adaptive permutation matrix.\label{fig:conv}}
\end{figure*}

The proposed privacy-preserving framework consists of an image encryption scheme and a modified ConvMixer network~\cite{p4} with an adaptive permutation matrix. Figure~\ref{fig:conv} depicts the proposed framework. A user encrypts training and testing images using the proposed image encryption scheme with a secret key and then sends the encrypted images to the cloud provider. The provider trains and tests the modified ConvMixer using these encrypted images. Data privacy can be protected in this process because the cloud provider has neither visual information of plain images nor the key.
\subsection{Image encryption method}

\begin{figure}[htbp]
	
	\begin{minipage}[b]{.48\linewidth}
		\centering
		\centerline{\includegraphics[width=3.5cm]{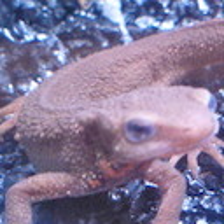}}
		\centerline{(a) Plain}\medskip
	\end{minipage}
	\hfill
	\begin{minipage}[b]{0.48\linewidth}
		\centering
		\centerline{\includegraphics[width=3.5cm]{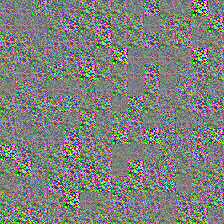}}
		\centerline{(b) Encrypted}\medskip
	\end{minipage}
	\caption{Example of plain image and encrypted image.\label{fig:example}}

\end{figure}

The proposed encryption method considers the property of the patch embedding structure in ConvMixer where the patch size is $M\times M$.
The procedure of the propose encryption method is as follows.
\begin{enumerate}
\item	Divide an 8-bit RGB image into blocks with a block size of $M \times M$.
\item	Permutate the divided blocks randomly with a secret key $ K_{1} $.
\item	Perform pixel shuffling in every block with a secret key $ K_{2} $, where $ K_{2} $ is commonly used in all blocks.
\item	Apply negative-positive transformation to each pixel in each block by using a secret key $ K_{3} $, where $ K_{3} $ is commonly used in all blocks.
\item	Concatenate all the blocks to produce an encrypted image.
\end{enumerate}
Figure~\ref{fig:example} shows an example of such an encrypted image.

\subsection{ConvMixer with adaptive permutation matrix}
\begin{figure}[htbp]
	\centerline{\includegraphics[width=9cm]{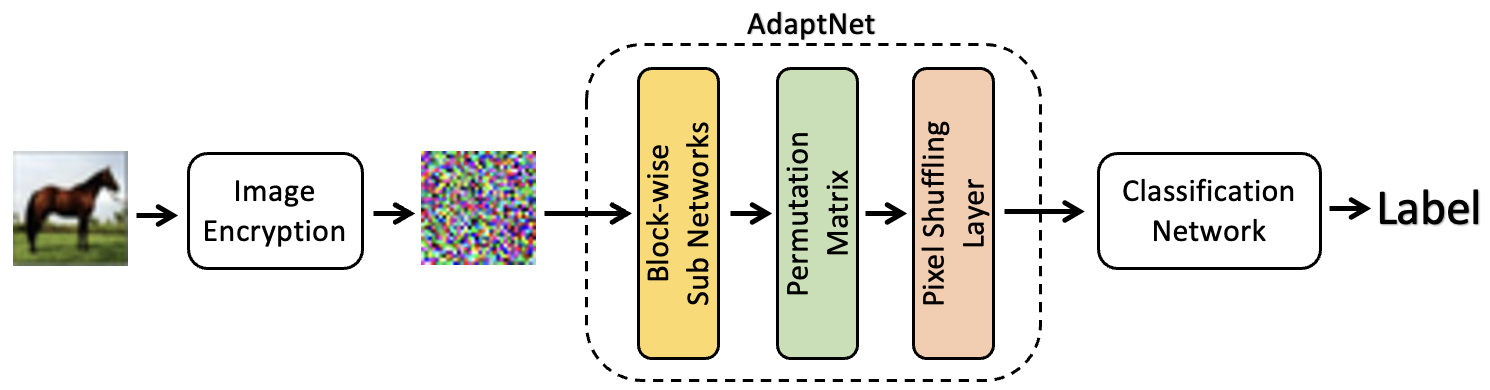}}
	\caption{Framework of conventional method with adaptation network.\label{fig:adaptnet}}
\end{figure}

Figure~\ref{fig:adaptnet} shows the framework of the conventional method with an adaptation network (ELE), where the adaptation network consists of block-wise sub-networks, a permutation matrix and a pixel shuffling layer. In contrast, Figure~\ref{fig:conv} shows the proposed one, in which a trainable permutation matrix is added after patch embedding, and then a resulting embedding is used as an input to ConvMixer Layers. The loss function used for the proposed method is given by

\begin{equation}
	L=L_{CE}+\lambda L_{U},
\end{equation}
where $L_{CE}$ is the cross-entropy loss, $L_{U}$ is the penalty for the adaptive permutation matrix introduced in~\cite{p3}, and $\lambda$ is a hyperparameter.

The proposed ConvMixer has two properties:
\begin{enumerate}
\item ConvMixer is carried out on the basis of a patch-based operation, so the proposed method can reduce the influence of block-wise encryption without block-wise sub-networks in Fig.~\ref{fig:adaptnet}.
\item Patch embedding can be adapted to pixel shuffling because pixel shuffling can be expressed as an invertible linear transformation, which is also learnable. 
\end{enumerate}

Therefore, the structure in Fig.~\ref{fig:conv} does not need to include block-wise sub-networks and pixel shuffling layers. An adaptive permutation matrix aims to reduce the influence of block permutation.

\subsection{Analysis of number of parameters in the model}
\label{nump}
In this section, we will analyze the number of parameters in each model under different image sizes.
\subsubsection{Classifier with adaptation network}
Conventional methods such as ELE need the combined use of an adaptation network and a classifier for improving the classification performance (see Fig.~\ref{fig:adaptnet}). In the adaptation network, sub-networks transform each block using a convolutional layer (with $3 \times output\_channel \times kernel\_size^{2}$ parameters) and a BatchNorm2d (with $2 \times output\_channel$ parameters) separately, and then the results are integrated and multiplied by a permutation matrix ($n \times n$ parameters).
Let $output\_channel$ (hidden size) be $h$ and $kernel\_size$ be $k$.
When an 8-bit RGB image is segmented into blocks with a block-size of $M$, there are $n$ blocks.
Note that the sub-networks in the adaptation network are intended to reduce the influence of encryption, thus $kernel\_size$ and block size $M$ are the same.
The total number of trainable parameters in the ELE~\cite{p3} is given as
 
\begin{equation}
\label{e1}
	\begin{aligned}
	N_{ELE}&=N_{AdaptNet}+N_{classifier}
	                   \\&=N_{sub-networks}+N_{matrix}+N_{classifier}
                       \\&=n(3\cdot h \cdot M^{2} + 2\cdot h)+n^{2}+N_{classifier}.      
	 \end{aligned}
\end{equation}

\subsubsection{ConvMixer with adaptive permutation matrix}
Unlike the ELE, the proposed method adds a permutation matrix only to the ConvMixer.
The number of parameters in the ConvMixer is given as in the original paper~\cite{p4},

\begin{equation}
  N_{ConvMixer} = h[d(k^2 + h + 6) + 3M^2 + n_{\text{classes}} + 3] + n_{\text{classes}},
\end{equation}
where $h$ is hidden size, $d$ is depth, $k$ is kernel size, and $n_{\text{classes}}$ is number of classes.
Note that we use the block size $M$ as a patch size in the ConvMixer.
The total number of parameters for the modified ConvMixer is given as

\begin{equation}
\label{e2}
	\begin{aligned}
	N_{Proposed}&=N_{ConvMixer}+N_{matrix}	
	                     \\&=N_{ConvMixer} + n^2.
    \end{aligned}
\end{equation}



\robustify\bfseries
\sisetup{table-parse-only,detect-weight=true,detect-inline-weight=text,round-mode=places,round-precision=2}
\begin{table*}[htbp]
	\caption{Classification accuracy  (\SI{}{\percent}) on CIFAR10 dataset of proposed and conventional privacy-preserving image classification methods. Best results are in bold.\label{tab:results}}
	\centering
	\resizebox{1.8\columnwidth}{!}{
		\begin{tabular}{lccccc}
			\toprule
			\multirow{2}{*}{\bfseries Encryption} & \multirow{2}{*}{\bfseries Network} & \multirow{2}{*}{\bfseries Image size} & \multirow{2}{*}{\bfseries Block-size} & \bfseries \# Parameters& \multirow{2}{*}{\bfseries Accuracy(\SI{}{\percent})}\\
			& &&& {$\approx(\times 10^{6}) $} &\\
			\midrule
			
			{EtC~\cite{p3,p7}} & AdaptNet+ShakeDrop & $ 3 \times 32 \times 32$ & $4 \times 4$ & 29.31  &89.09\\
			{ELE~\cite{p3}} & AdaptNet+ShakeDrop  &  $ 3\times 32 \times 32$ & $4 \times 4$ & 29.31 &83.06\\
			{Proposed} & ConvMixer & $3 \times 224 \times 224 $ & $16 \times 16$ & {\bfseries \num{5.31}} & 89.14 \\
			{Proposed} & ConvMixer$^{\dagger}$ & $ 3 \times 224 \times 224$ & $16 \times 16$ &{\bfseries \num{5.35}} &{\bfseries \num{92.65}} \\
			\midrule
			Plain & ShakeDrop  & $ 3\times 32 \times 32$ & N/A & 28.49 & 96.70\\
			Plain & ConvMixer  & $ 3\times 224 \times 224$ & N/A & 5.31 &96.80\\
			\bottomrule
			\multicolumn{4}{l}{$^{\dagger}$ ConvMixer with an adaptive permutation matrix (proposed).}
		\end{tabular}
	}
\end{table*}

\subsubsection{Comparison under different image sizes}
\begin{figure}[htbp]
	\centerline{\includegraphics[width=9cm]{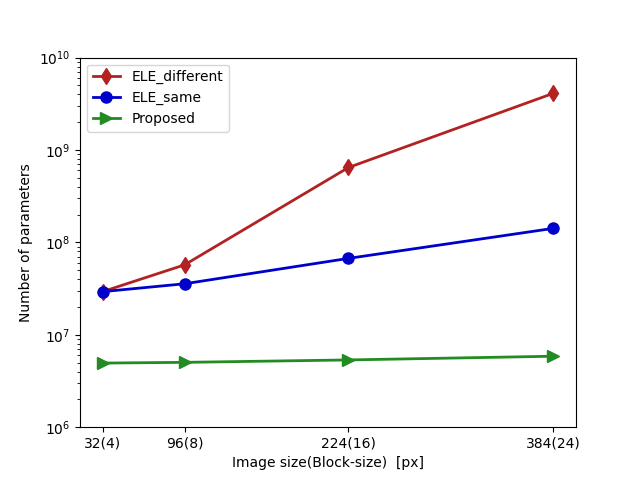}}
	\caption{Number of parameters for ELE~\cite{p3} and proposed method under different image sizes.\label{fig:params}}
\end{figure}

We observe the number of parameters in the ELE and the proposed method when using different image sizes.
Figure~\ref{fig:params} shows the graph of the number of parameters calculated as in Eq.~\eqref{e1} and \eqref{e2} versus image sizes.
For the adaptation network of ELE, when the size of input images becomes larger, using the same hidden size $h$ (denoted as ELE\_same) for convolutional layers in the sub-networks will lead to output representation with smaller number of channels.
This might degrade the performance of the classifier.
Using a larger hidden size $h$ (denoted as ELE\_different) can increase the number of channels in the output representation, but also increases the number of parameters in the adaptation network drastically.
In contrast, the proposed method does not increase the number of parameters significantly even when large image sizes are used.

\section{Experiment Results}
\label{Exp}

We conducted image classification experiments on the CIFAR-10 dataset~\cite{p6}.
Images were resized to $224 \times 224$ for the proposed encryption and the original image size ($32\times 32$) was used for conventional encryption methods: EtC~\cite{p7} and ELE~\cite{p3} with a block size of $4\times 4$. 
The configurations of ConvMixer were: a kernel size of 9, a depth of 16, and a hidden size of 512. The patch size of the ConvMixer was always the same as the block-size in the proposed encryption. We used the training settings from~\cite{p4} except for the training epochs. We trained the ConvMixer models for 300 epochs for plain images, and 400 epochs for encrypted images. In addition, hyperparameter $\lambda$ in the loss function was set to 0.0001.

Table~\ref{tab:results} shows the image classification performance of the proposed method compared with state-of-the-art methods with an adaptation network such as EtC~\cite{p7} and ELE~\cite{p3}.
The ConvMixer model with an adaptive permutation matrix achieved the highest classification accuracy for images encrypted by the proposed encryption method.
In contrast, without the permutation matrix, the ConvMixer model decreased the accuracy by approximately \SI{3}{\percent}.
Similarly, conventional methods with an adaptation network also dropped the accuracy.
All in all, ConvMixer models provided higher accuracy for both plain and encrypted images.

Table~\ref{tab:results} also shows the number of parameters of each model calculated by Eq.~\eqref{e1} and \eqref{e2}. ConvMixer models are significantly smaller than convolutional networks such as ShakeDrop. 
As shown in Table~\ref{tab:results}, the ShakeDrop with an adaptation network used in~\cite{p7, p3} are extremely large even for a smaller image size ($32\times 32$) and a block size $4\times 4$.
The number of parameters in the adaptation network will further be increased if a large image size is used as we explained in section \ref{nump}.
Therefore, the proposed privacy-preserving image classification with the ConvMixer provides a higher classification accuracy with less number of parameters.

\section{Conclusion}

In this paper, we proposed a novel privacy-preserving image classification method that uses the ConvMixer network with an adaptive permutation matrix. An image encryption method was also proposed for training ConvMixer models and testing. In an experiment, the proposed method was demonstrated to outperform conventional methods in terms of classification accuracy and the number of parameters. 

\bibliographystyle{IEEEtran}
\bibliography{IEEEabrv,refs}
\end{document}